\documentclass{article}

\usepackage[nonatbib,final]{bsw_2022}

\usepackage[utf8]{inputenc} 
\usepackage[T1]{fontenc}    
\usepackage{hyperref}       
\usepackage{url}            
\usepackage{booktabs}       
\usepackage{amsfonts}       
\usepackage{nicefrac}       
\usepackage{microtype}      
\usepackage{xcolor}   

\usepackage[backend=biber,sorting=none]{biblatex}
\usepackage{tikz}           
\usepackage{amsmath}

\makeatletter
\def\blfootnote{\xdef\@thefnmark{}\@footnotetext}
\makeatother

\title{Improving Neural Predictivity in the Visual Cortex with
Gated Recurrent Connections}
\author{%
  Simone Azeglio\\
  Hearing Institute\\
  Institut Pasteur\\
  Paris, FR 75012 \\
  \texttt{simone.azeglio@pasteur.fr} \\
  \And
  Simone Poetto\\
  Nicolaus Copernicus University\\
  Torun, Poland\\
  ISI Foundation\\
  Turin, Italy\\
  \texttt{poets@doktorant.umk.pl}
  \And
  Luca Savant Aira \\
  Politecnico di Torino\\
  Torino, IT 10129\\
  \texttt{luca.savantaira@studenti.polito.it}
  \And
  Marco Nurisso \\
  Politecnico di Torino\\
  Torino, IT 10129\\
  \texttt{marco.nurisso@studenti.polito.it}
}

\begin{document}

\maketitle
\begin{abstract}
Computational models of vision have traditionally been developed in a bottom-up fashion, by hierarchically composing a series of straightforward operations - i.e. convolution and pooling. The aim was to emulate simple and complex cells in the visual cortex and resulted in the introduction of deep convolutional neural networks (CNNs).
Nevertheless, evidence obtained with recent neuronal recording techniques suggests that the nature of the computations carried out in the ventral visual stream is not completely captured by current deep CNN models. To fill the gap between the ventral visual stream and deep models, several benchmarks have been designed and organized into the Brain-Score platform, granting a way to perform multi-layer (V1, V2, V4, IT) and behavioral comparisons between the two counterparts. In our work, we shift the focus to architectures that take into account lateral recurrent connections, a ubiquitous feature of the ventral visual stream, to devise adaptive receptive fields. Through recurrent connections, the input’s long-range spatial dependencies can be captured in a local multi-step fashion and, as introduced with Gated Recurrent CNNs (GRCNN), the unbounded expansion of the neuron’s receptive fields can be modulated through the use of gates. To increase the robustness of our approach and the biological fidelity of the activations, we employ specific data augmentation techniques in line with several of the scoring benchmarks. We find that forcing some form of invariance, through heuristics, resulted in better neural predictivity.

\end{abstract}
\section{Introduction}

As suggested by abundant physiological evidence, recurrent circuits are ubiquitous in the visual cortex \cite{deco2004role, zhu2013visual, kubilius2018cornet} and in several areas of the mammalian brain \cite{chance1999complex}. In visual layers, the effect of lateral recurrent connections is believed to contribute to receptive fields adaptation \cite{nelson1978orientation,jones2002spatial,cavanaugh2002nature,series2003silent}.
More generally, there are many indications of the capability of the brain to modulate the processing of visual signals on the basis of their context \cite{allman1985stimulus,croner1999seeing,albright2002contextual}. 
Classical CNNs, despite showing many similarities with the ventral visual stream, lack the ability of context modulation. Trying to incorporate recurrent connections in a CNN architecture poses a problem because standard Recurrent Neural Networks (RNNs) are designed to process time varying sequences of inputs, while computer vision models deal with static inputs. To both circumvent this limit and try to design a more biologically plausible architecture, the basic idea is to introduce a form of recursion across neurons of the same layer. In this way it is possible to avoid the problem of static inputs, but still give to each neuron in a layer information about the activity of the surrounding neurons, allowing them to receive information about a larger part of the image.
Similar solutions have been proposed in different forms by many authors, often with the aim of better simulating the visual system \cite{liang2015recurrent,kubilius2018cornet,spoerer2020recurrent}.
It is worth noticing that the computational graphs of these models, when unfolded in time, look like pure feedforward hierarchies of operations enriched with a number of skip connections. This results in  architectures that are very similar to residual networks \cite{he2016deep}.
To achieve even better biological adherence, it is useful to add gates in between the recursive computations, as proposed by \cite{wang2021convolutional}. In this way the neurons' receptive fields are explicitly modeled and modulated by the gates.

\section{Model}

\begin{figure}[!ht]
    \centering
    \input{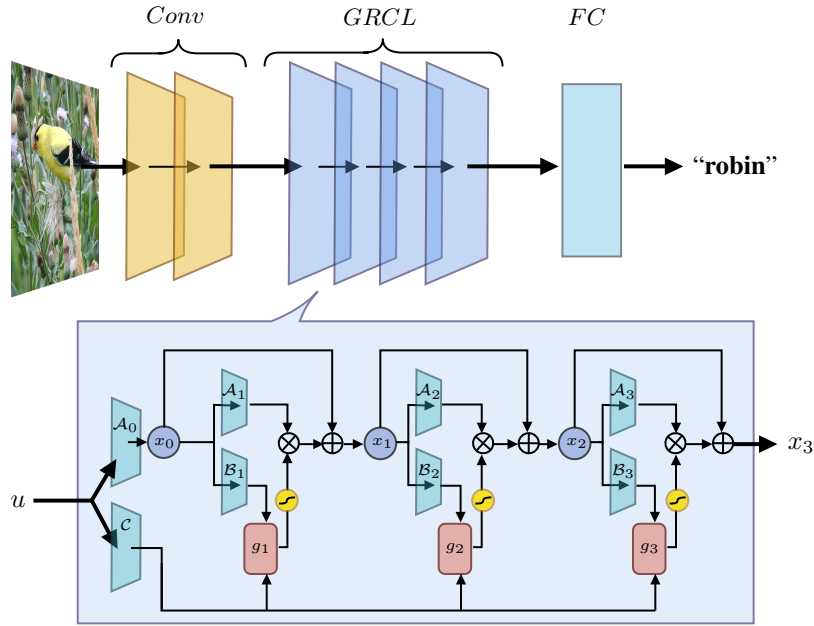}
    \caption{\textbf{Top} GRCNN architecture trained on the ImageNet dataset. It is composed by 2 Convolutional layers, 4 GRCL blocks and a Fully Connected layer which is employed for classification.
    \textbf{Bottom} Schematic representation of a GRCL block.
    Note that if we unfold the computational graph through time, we end up with something that is very similar to a ResNet}.
    \label{fig:GRCNN}
\end{figure}

The introduction of gates is a distinctive feature of GRCNN and is motivated in \cite{wang2021convolutional}. The role of gates can be intuitively and qualitatively understood as an extra layer of computation resembling an attention mechanism
\cite{guo2021attention}. Gates are designed to give an output between 0 and 1 that multiplies (pointwise) the activations in the recurrent convolutional layer. This means that, during training, the set of weights associated with each gate evolves in such a way as to give the network the capability to notice which parts of the image are relevant.
In more detail, a GRCNN is composed by a feedforward sequence of blocks called gated recurrent convolutional layers (GRCL).
Every GRCL block computes a recursion on its inputs, and, between each recurrent operation, the gate system modulates the effective amount of forwarded information - see \textit{Figure} \ref{fig:GRCNN} \textbf{Bottom}.

The equations describing the computations inside the GRCL are the following:

\begin{equation}
\begin{cases}
x_0 = \mathcal{A}_0(u)\\
g_t = \mathcal{B}_t(x_{t-1})+\mathcal{C}(u), &t = 1,2,\dots,T\\
x_t = x_{t-1} + \sigma(g_t) \odot \mathcal{A}_t(x_{t-1}), &t = 1,2,\dots,T
\end{cases}    
\end{equation}

where
$u$ is the input, $\mathcal{A}_t, \mathcal{B}_t, \mathcal{C}$ are convolutional operators with nonlinear activation functions (ReLU) and batch normalization, $\sigma$ is the logistic sigmoid function and $\odot$ is the Hadamard product (elementwise multiplication operator). We emphasize the variables $x_t$ and $g_t$ which represent respectively the recurrent state variables and gate activations. The third equation clearly shows how the absence of gates would lead GRCL to be a standard \textit{ResNetT}.





The complete model architecture is obtained by stacking two initial convolutional blocks, four GRCLs with $T=3$, and a final readout unit (see \textit{Figure \ref{fig:GRCNN}} \textbf{Top}). The architecture's  number of parameters is comparable with ResNet50. 
For further technical details, we refer the reader to the original GRCNN paper \cite{wang2021convolutional}.

\section{Augmentation \& Regularization}

One of the main limitations in devising a deep model that predicts the neural activity of the ventral visual stream is the training procedure. Usually, vision models are trained on the ImageNet dataset \cite{deng2009imagenet}, which in its original formulation is related to a multiclass classification problem. On the other side, Brain-Score's benchmarks include several tasks \cite{SchrimpfKubilius2018BrainScore,Schrimpf2020integrative} which are not necessarily related to classification.
In our work, we try to alleviate this problem by employing and designing augmentation and regularization strategies, inspired by some of the evaluation benchmarks. In particular, given their importance in the overall scoring, we decided to focus on behavior - more details can be found in \cite{rajalingham2018large} -, V1 and V2 tasks based on \cite{freeman2013functional}. 

Given the spirit of the behavioral benchmark, we took advantage of \textit{CutMix} \cite{yun2019cutmix}, an augmentation strategy that facilitates the recognition of different objects from partial views in a single image. More specifically, in CutMix, patches are cut and pasted among training images while, at the same time, the corresponding labels are linearly combined, resulting in: $\Tilde{x} =  \mathbf{M} \odot x_{A} + (\mathbf{1}- \mathbf{M}) \odot x_{B}$ and $\Tilde{y} =  \lambda y_A + (1-\lambda) y_B $, where $\mathbf{M} \in \{0,1\}^{W \times H}$ denotes a binary mask indicating where to drop out and fill in from two images, $\mathbf{1}$ is a binary mask filled with ones and $\lambda$ is sampled from the uniform distribution $U(0,1)$. 

In parallel, to enhance the robustness of learned representations, we employed \textit{AugMix} \cite{hendrycks2019augmix}, which allows our model to explore the semantically equivalent input space around an image. Briefly, AugMix consists in combining simple augmentation operations - e.g. translation, rotation, shear - together with a consistency loss. Augmentations are sampled stochastically and concatenated while a consistent embedding around the input image is enforced by using the Jensen-Shannon divergence as a loss. 

Lastly, we introduced a regularization term in a similar vein to \cite{li2019learning}, with the idea that behavioral traits cannot be fully described in terms of a scalar metric - e.g. accuracy - but need to be conceived in terms of higher-order descriptors such as reconstructing a confusion matrix, in order to force a network to fail in the same way as a primate would. We took as a reference human performances on 11 superclasses (i.e. groups of classes) of ImageNet \cite{tsipras2020imagenet}, projected down the predictions of our model - 1000 dimensional vector - onto the 11 superclasses and imposed an additional cross entropy loss on this term, to quantify the distance between our model and human performances. 

To improve scores on V1 and V2 benchmarks, we opted for an augmentation technique largely inspired by \cite{freeman2013functional}.
Given that V2 neurons are particularly sensible to textures while V1 neurons respond in a similar way to a texture and its phase randomized counterpart, we considered a texture dataset composed of images coming from the following \cite{Kylberg2011c,lazebnik2005sparse,Fritz2004THEKD}, and we extended it by generating each texture's phase randomized counterpart, named \textit{noise} \footnote{We open-sourced generated Textures and Noise dataset at \href{https://github.com/sazio/TexturesNoiseDataset}{https://github.com/sazio/TexturesNoiseDataset} }. After that, we fine-tuned our model's first GRCL block - roughly corresponding to V1 - by freezing all the other layers and randomly blending input images separately with textures and noise. Later on, we fine-tuned the second GRCL block in the same way, but by only blending input images with textures. In this way we tried to emulate how specific layers in the visual cortex respond to different stimuli, and we got improvements in both V1 and V2 with respect to the baseline.

\section{Results and conclusions}
Introducing receptive fields modulation through a gated recurrent mechanism is beneficial in terms of neural predictivity in the visual cortex. The baseline GRCNN model (see GRCNN55 and GRCNN109 in \ref{Brainscore_table}), regularly trained on Imagenet without augmentation, shows promising results in several benchmarks. As shown in table \ref{Brainscore_table}, by introducing the previously mentioned regularization and augmentation techniques, scores improve on specific benchmarks, as well as on average.

In the current state, further work is needed to get a better sense of such preliminary results \footnote{We open-sourced the code for our experiments at \href{https://github.com/sazio/brainscore2022}{https://github.com/sazio/brainscore2022}}. In this regard, we have implemented several variants of the baseline model, including an integration of the VOneBlock \cite{dapello2020simulating} as a substitute of the first convolutional block in the GRCNN. With more computational power we are planning to train this combined architecture from scratch. Ultimately, another interesting perspective for further exploration is related to how CNNs learn textures and not shapes \cite{geirhos2018imagenet}, which is ultimately analogous to our augmentation strategy for V1 and V2.


\section{Acknowledgment}

All the authors wish to show their appreciation to MLJC\footnote{\href{https://www.mljc.it/}{https://www.mljc.it/}} for providing them with the essential computational power (Nvidia RTX 2080 Ti) for finetuning their models. S.A. would like to thank Arianna Di Bernardo and Sophie Bagur for useful and constructive discussions as well as COSYNE 2022 organizers for providing him with a Travel Grant to take part in the conference. S.P. would like to thank ISI Foundation\footnote{\href{https://www.isi.it/en/home}{https://www.isi.it}} for the hospitality while performing this work and for the computational resources, the Centre of Informatics Tricity Academic Supercomputer \& Network in Gdansk for additional computational resources that will be crucial for further development, Giovanni Petri and Karolina Finc for the kind supervision, and Alan Perotti, Maxime Lucas, Antonio Leitao for the helpful discussions. S.P. would also like to thank the Polish National Science Foundation grant UMO-2016/20/W/NZ4/00354 for partially funding this work.

\printbibliography


\clearpage
\section*{Appendix}

\begin{table}[!hb]
\centering
\begin{tabular}{|c||c|c|c|c|c|c|}
    \hline
    Name & Average & V1 & V2 & V4 & IT & Behavior\\
    \hline \hline
     GRCNN55 Behavior & .463 & .509 & .303 & .482 & .467 & .554 \\
     GRCNN55 & .462 & .525 & .306 & .481 & .479 & .520 \\
     GRCNN109 & .461 & .520 & .328 & .475 & .464 & .521 \\
     GRCNN55 V1-V2 & .458 & .535 & .314 & .486 & .481 & .473 \\
     ResNet50 & .427 & .497 & .264 & .465 & .475 & .432 \\
     AlexNet & .424 & .508 & .353 & .443 & .447 & .370 \\
    \hline
\end{tabular}
\caption{Brainscore benchmarks results, the number in the name of each architecture represents the depth}
\label{Brainscore_table}
\end{table}

\end{document}